\documentclass{article}

\PassOptionsToPackage{numbers, compress}{natbib}

\usepackage[preprint]{neurips_2023}

\usepackage[utf8]{inputenc} %
\usepackage[T1]{fontenc}    %
\usepackage{hyperref}       %
\usepackage{url}            %
\usepackage{booktabs}       %
\usepackage{amsfonts}       %
\usepackage{nicefrac}       %
\usepackage{microtype}      %
\usepackage{xcolor}         %
\usepackage{xspace}
\usepackage{todonotes}
\usepackage{caption}
\usepackage{subcaption}
\usepackage{enumitem} %
\usepackage{adjustbox} %
\usepackage{array} %

\usepackage{booktabs}
\usepackage{multirow}
\newcommand{\chk}{\checkmark}

\newcommand{\eg}{\textit{e.g.}}

\newcommand{\pkgName}{Event Stream GPT\xspace}
\newcommand{\pkgAbbr}{ESGPT\xspace}

\newcommand{\mr}[1]{\multirow{2}{*}{#1}}

\usepackage{amsmath}

\renewcommand{\vec}{\boldsymbol}

\newcommand{\block}[1]{%
  \raisebox{\dimexpr(\fontcharht\font`X-1em)/2}{\rule{1em}{#1\dimexpr1em/8}}%
}

\newcolumntype{R}[2]{%
    >{\adjustbox{angle=#1,lap=\width-(#2)}\bgroup}%
    l%
    <{\egroup}%
}
\newcommand*\rot{\multicolumn{1}{R{45}{1em}}}%
\newcommand{\rbox}{\rot}

\DeclareUnicodeCharacter{2581}{\block{1}}
\DeclareUnicodeCharacter{2582}{\block{2}}
\DeclareUnicodeCharacter{2583}{\block{3}}
\DeclareUnicodeCharacter{2584}{\block{4}}
\DeclareUnicodeCharacter{2585}{\block{5}}
\DeclareUnicodeCharacter{2586}{\block{6}}
\DeclareUnicodeCharacter{2587}{\block{7}}
\DeclareUnicodeCharacter{2588}{\block{8}}

\usepackage{listings}
\usepackage{xcolor}
\usepackage{caption}
\usepackage{float}
\newcommand\YAMLcolonstyle{\color{red}\mdseries}
\newcommand\YAMLkeystyle{\color{black}\bfseries}
\newcommand\YAMLvaluestyle{\color{blue}\mdseries}

\makeatletter

\newcommand\language@yaml{yaml}

\expandafter\expandafter\expandafter\lstdefinelanguage
\expandafter{\language@yaml}
{
  keywords={true,false,null,y,n},
  keywordstyle=\color{darkgray}\bfseries,
  basicstyle=\YAMLkeystyle,                                 %
  sensitive=false,
  comment=[l]{\#},
  morecomment=[s]{/*}{*/},
  commentstyle=\color{purple}\ttfamily,
  stringstyle=\YAMLvaluestyle\ttfamily,
  moredelim=[l][\color{orange}]{\&},
  moredelim=[l][\color{magenta}]{*},
  moredelim=**[il][\YAMLcolonstyle{:}\YAMLvaluestyle]{:},   %
  morestring=[b]',
  morestring=[b]",
  literate =    {---}{{\ProcessThreeDashes}}3
                {>}{{\textcolor{red}\textgreater}}1     
                {|}{{\textcolor{red}\textbar}}1 
                {\ -\ }{{\mdseries\ -\ }}3,
}

\lst@AddToHook{EveryLine}{\ifx\lst@language\language@yaml\YAMLkeystyle\fi}
\makeatother

\usepackage[]{roboto-mono}
\definecolor{textblue}{rgb}{.2,.2,.7}
\definecolor{textred}{rgb}{0.54,0,0}
\definecolor{textgreen}{rgb}{0,0.43,0}

\lstdefinestyle{configs}{
    language=yaml, 
    numbers=left, 
    numberstyle=\tiny, 
    stepnumber=1,
    numbersep=5pt, 
    tabsize=4,
    basicstyle=\ttfamily,
    frame=none,                    
    columns=fullflexible,
    keepspaces=true,
    xleftmargin=\parindent,
    showstringspaces=false
}
\lstdefinestyle{code}{
    language=Python, 
    numbers=left, 
    numberstyle=\tiny, 
    stepnumber=1,
    numbersep=5pt, 
    tabsize=4,
    basicstyle=\small\ttfamily,
    keywordstyle=\color{textblue},
    commentstyle=\color{textred},   
    stringstyle=\color{textgreen},
    frame=none,                    
    columns=fullflexible,
    keepspaces=true,
    xleftmargin=\parindent,
    showstringspaces=false
}

\title{\pkgName: A Data Pre-processing and Modeling Library for Generative, Pre-trained Transformers over Continuous-time Sequences of Complex Events}

\author{%
  Matthew B. A. McDermott \\
  Harvard University\\
  \texttt{matthew\_mcdermott@hms.harvard.edu} \\
  \And 
  Bret Nestor \\
  Massachusetts Institute of Technology \\
  \And
  Peniel Argaw \\
  Harvard University \\
  \And 
  Isaac Kohane \\
  Harvard Univiersity \\
}

\begin{document}

\maketitle

\begin{abstract}
Generative, pre-trained transformers (GPTs, \textit{a.k.a.} "Foundation Models") have reshaped natural language processing (NLP) through their versatility in diverse downstream tasks. However, their potential extends far beyond NLP. This paper provides a software utility to help realize this potential, extending the applicability of GPTs to continuous-time sequences of complex events with internal dependencies, such as medical record datasets. Despite their potential, the adoption of foundation models in these domains has been hampered by the lack of suitable tools for model construction and evaluation. To bridge this gap, we introduce \pkgName (\pkgAbbr), an open-source library designed to streamline the end-to-end process for building GPTs for continuous-time event sequences. \pkgAbbr allows users to (1) build flexible, foundation-model scale input datasets by specifying only a minimal configuration file, (2) leverage a Hugging Face compatible modeling API for GPTs over this modality that incorporates intra-event causal dependency structures and autoregressive generation capabilities, and (3) evaluate models via standardized processes that can assess few and \emph{even zero-shot} performance of pre-trained models on user-specified fine-tuning tasks.
\end{abstract}

\section{Introduction}
The unprecedented performance of large language models (LLMs) in natural language processing (NLP) has ushered in a new paradigm in machine learning. Instead of being dominated by single-task, supervised learning models, this paradigm is characterized by general-purpose, pre-trained ``foundation models.'' These models, exemplified by the ``Generative, Pre-trained Transformer'' (GPT) architecture, deliver state-of-the-art performance on diverse downstream tasks in a highly data-efficient manner~\cite{gpt4}.
The success of GPT models over natural language data is driven in part by the fact that the generative, language modeling forecasting task is ``universal'' across NLP downstream tasks---meaning any natural language task can be recast as a language modeling task through prompting. However, this property is not unique to NLP, and GPT-equivalent architectures may hold significant promise in other domains as well. In particular, this work explores the application of these models to the domain of continuous-time sequences of complex, multi-modal events (event streams), specifically electronic health record (EHR) datasets. %

Two critical barriers inhibit the application of foundation model research to modalities like EHR data.
First, these data modalities do not come in a single, unified format, and compelling datasets are often private and unshareable. This hampers foundation model research by requiring dataset-specific pre-processing pipelines and hindering one's ability to assess the model generalizability.
Second, modeling continuous-time sequences of complex events is more challenging than modeling ordinal sequences of tokens, as GPTs over these modalities must account for non-ordinal inputs, complex, multi-modal emission distributions, and intra-event causal relationships.

In this paper, we address these gaps with \pkgName (\pkgAbbr), an open source software package\footnote{Code provided in Supplementary Material}, API, and evaluation utility for foundation models over event stream data. \pkgAbbr can represent diverse datasets across various sources in a unified manner, pre-process very large datasets extremely quickly through its use of the Polars library~\cite{polars_software}, and can Hyperparameter tune, pre-train, fine-tune, and perform zero-shot evaluation for foundation models through a Hugging Face compatible API.
In the rest of this work, we will demonstrate the unique value \pkgName offers through the lens of a real-world, working example: building foundation models over the MIMIC-IV electronic health record dataset~\cite{mimiciv} (Section~\ref{sec:working_example}). We will walk through all aspects of the pipeline one would follow to pursue this research, revealing in each how \pkgAbbr fills a critical gap in existing tools across data pre-processing (Section~\ref{sec:data}), model configuration and running (Section~\ref{sec:model}), and evaluation (Section~\ref{sec:evaluation}).
Finally, we will close with related work and concluding thoughts.

\section{Working Example Problem Setup} \label{sec:working_example}
\paragraph{Source Data}
The MIMIC-IV dataset~\cite{mimiciv} is a publicly available dataset consisting of the EHR data for all adult patients who were admitted to the emergency department or an intensive care unit (ICU) at Beth Israel Deaconess Medical Center between 2008 and 2019. This dataset contains approximately 300,000 patients and consists of numerous modalities of health data, including diagnoses, laboratory test results, medications, in and out of hospital mortality, and many others, all localized continuously in time. For our running example, we will define the internal covariates of each event to include the set of modeling targets in Table~\ref{tab:included_modalities}.

\paragraph{Modeling Task}
Our sample modeling challenge is to build a GPT model over the continuous-time, complex event stream data contained in MIMIC-IV. This can also be seen as a multi-variate marked temporal point process. In particular, given a sequence of complex events $\vec x_1, \ldots, \vec x_N$ (where each event $\vec x_i$ is a collection of occurrences of the selected covariates in Table~\ref{tab:included_modalities}) which occur at continuous times $t_1, \ldots, t_N$, we wish to model the following probability distribution:
\begin{align*}
    p(t_i, \vec x_i | \underbrace{(t_1, \vec x_1), \ldots, (t_{i-1}, \vec x_{i-1})}_{\vec h_{i-1}})
\end{align*}
We will realize this through a transformer neural network architecture parametrized by $\vec \theta$, such that $f_{\vec \theta} (t_i, \vec x_i, \vec h_{i-1}) = p(t_i, \vec x_i | \vec h_{i-1})$. Note that, unlike other GPT modeling settings, here it may be the case that internal covariates of each event $\vec x_i$ have internal causal dependencies. For example, if $\vec x_i^{(j)}$ is used to denote the $j$th internal covariate of event $i$, then $p(\vec x_i | \vec h_{i-1}, t_i) \neq \prod_{j} p(\vec x_i^{(j)} | \vec h_{i-1}, t_i)$. Any full generative model will therefore need to account for these internal causal dependencies.

\begin{table}[!ht]
    \centering
    \begin{tabular}{lll} 
        Admission/Discharge   & Demographics   & Measurements/Actions \\ \midrule
        Admission Type        & Language       & Laboratory Tests$^*$ \\
        Admission Location    & Race$^\dagger$ & Procedures           \\
        ICU Care-unit         & Marital Status & Medications          \\
        Discharge Destination & Insurance Type & Infusions$^*$        \\
                              & Age            & Diagnoses            \\
                              & Time-of-day    & Patient Weight       \\ 
                              & Gender         &                      \\ 
    \end{tabular}    \caption{The modalities of data we include in our MIMIC-IV working example. Our generative model consists of continuous timeseries of partial observations of any of these events, and we task the model with predicting which of the times of these events, which modalities will be observed in those events, and what values they will take on. Included modalities are not observed uniformly across all events, in a non-random manner. 
    $*$: Also includes continuous regression components (laboratory test results and total infusion amount).}
    \label{tab:included_modalities}
\end{table}

\section{Data Extraction and Pre-processing}
\label{sec:data}

\subsection{The Problem}\label{subsec:data_problem}
To build a dataset suitable for performing our foundation modeling task outlined in Section~\ref{sec:working_example}, we perform the following steps:
\begin{enumerate}[nolistsep]
    \item Extract the raw data from its source (here, the MIMIC-IV postgresql database).
    \item Pre-process the data: learn outlier detection and normalization parameters for numerical variables, filter out infrequently measured variables or outliers, normalize data, etc.
    \item Join independent input datasets together into a format optimized for deep learning.
    \item Build a PyTorch dataset, dataloader, and embedding layer for this deep-learning friendly structure, such that we can use these data efficiently in downstream pipelines.
\end{enumerate}

\subsection{Current State-of-the-art}
If we wish to use current solutions to the data extraction problem, we must find an existing tool that either is fully specialized to our dataset (MIMIC-IV) and use-case (of which no such example exists), or is a general purpose data-extraction, pre-processing, and/or modeling pipeline that will ease our workflow. For such a general purpose tool to be useful, it must meet several critical criteria:
\begin{enumerate}[nolistsep]
    \item It must solve at least a majority of the challenges identified in Section~\ref{subsec:data_problem}.
    \item It must be flexible enough to handle not only the MIMIC-IV data, but also various other input datasets, schemas, and covariate choices.
    \item It must be designed for efficiency in mind -- both in data pre-processing and in the output deep-learning representations we produce, so that we can pre-train large models effectively.
    \item It must be actively maintained, well documented, and tested.
\end{enumerate}

Unfortunately, as shown in Table~\ref{tab:existing_pipelines}, existing pipelines for extracting Electronic Health Record (EHR) data for machine learning use do not fully meet these criteria. Only one pipeline meets even a meaningful subset of these critical desiderata, TemporAI~\cite{temporai}, as it has strong development practices, exposes a flexible input format, and meets our critical pre-processing feature goals; however, TemporAI does not support native extraction from underlying source files, and suffers in its efficiency, both of the core data pre-processing pipeline and in its output deep learning representation, raising concerns about its viability on larger datasets.

\begin{table}
    \centering \small
        \begin{tabular}{lcccccccccccccccc}
        \multirow{2}{*}{} & \rbox{Usable on General Datasets} & \rbox{Flexible Input Format} & \rbox{Performs Extraction from Source} & \rbox{Accepts Arbitrary Input Variables} & \rbox{Fits Vocabularies} & \rbox{Outlier Detection} & \rbox{Pre-set censoring/outlier removal} & \rbox{Normalization} & \rbox{Filters Infrequent Measurements} & \rbox{Efficiently Pre-processes Data using Modern Libraries$^\ddagger$} & \rbox{Produces Efficient Deep-Learning Representation$^*$} & \rbox{Produces PyTorch Datasets} & \rbox{Provides Embedding Layer} & \rbox{Actively Maintained$^\star$} & \rbox{Tested$^\dagger$} & \rbox{Comprehensive Documentation} \\ 
        \cmidrule(lr){2-5} \cmidrule(lr){6-9} \cmidrule(lr){10-12} \cmidrule(lr){13-17}  & \multicolumn{4}{c}{Input} & \multicolumn{4}{c}{Pre-processing} & \multicolumn{3}{c}{PyTorch} & \multicolumn{5}{c}{Dev}\\
        \midrule
        MIMIC-Extract~\cite{mimic_extract} &      &      & \chk &      &      &      & \chk & \chk & \chk &      &      &      &      &      &      &     \\
        Fiddle~\cite{fiddle}               & \chk &      &      &      & \chk &      &      & \chk & \chk &      &      &      &      &      &      &     \\
        MT-MIMIC~\cite{MTMIMIC}            &      &      &      &      & \chk &      &      & \chk &      &      &      &      &      &      &      &     \\
        Clairvoyance~\cite{clairvoyance}   & \chk & \chk &      & \chk & \chk & \chk &      & \chk & \chk &      &      & \chk & \chk &      &      & \chk\\
        OMOP-Learn~\cite{omoplearn}        & \chk &      & \chk &      &      &      &      & \chk &      &      &      & \chk &      &      &      &     \\
        TemporAI~\cite{temporai}           & \chk & \chk &      & \chk & \chk & \chk &      & \chk & \chk &      &      & \chk & \chk & \chk & \chk & \chk\\
        \pkgAbbr (Ours)                    & \chk & \chk & \chk & \chk & \chk & \chk & \chk & \chk & \chk & \chk & \chk & \chk & \chk & \chk & \chk & \chk\\
        \bottomrule \end{tabular}
        \caption{A collection of existing data pre-processing and/or modelling pipelines for temporal EHR data, categorized on whether or not they enable various features. $\star$ Active maintenance is defined by the existence of a commit between May 2022 and May 2023. $\dagger$ Tested is defined by the presence of a reasonably comprehensive, automated test suite. $\ddagger$ Efficient data pre-processing is defined by whether or not the library relies on Pandas~\cite{pandas_paper,pandas_software} for their dataframe manipulation, as opposed to more modern, faster systems such as Polars~\cite{polars_software} or DuckDB~\cite{duckdb}. $*$ Efficient deep-learning representation is defined by the extent to which sparsity is leveraged in their deep learning representation files; efficient outputs should only scale in size with the observations actually present in any given batch of data, not with the number of total features in the space, total possible sequence elements, etc.}
    \label{tab:existing_pipelines}
\end{table}

\subsection{Contributions of \pkgName}
\begin{figure}
    \centering
    \includegraphics[width=0.9\linewidth]{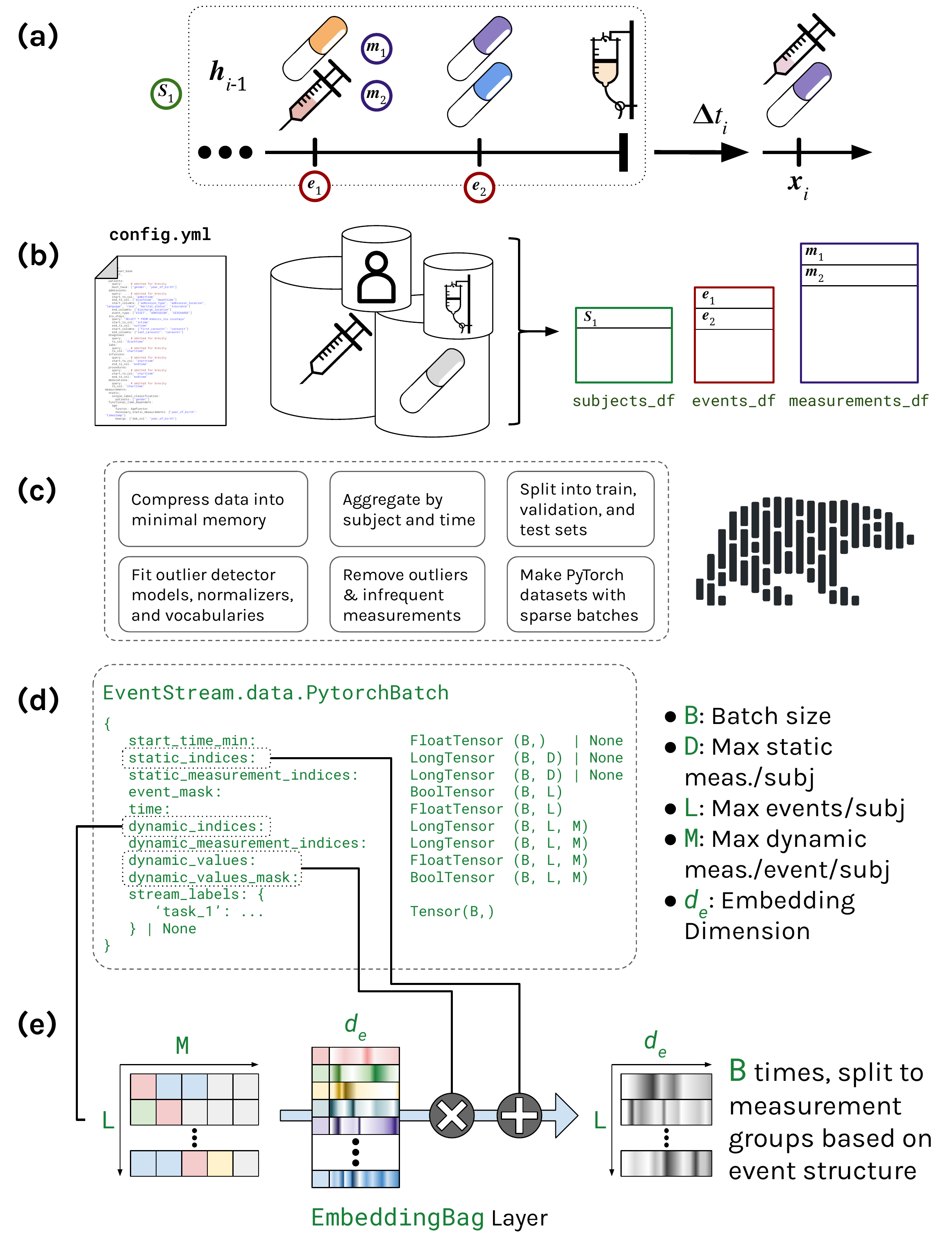}
    \caption{
        \pkgAbbr's end-to-end data pipeline, which spans raw data extraction from source all the way to production of a PyTorch dataset and pre-built embedder suitable for use in any deep learning pipeline.
        \textbf{(a)} An example of our data modality; for a single subject $\vec S_1$, the data consists of a series of events at continuous timestamps $\vec e_1, \vec e_2, \ldots$ (such as medical visits), with each event being composed of inter-related internal covariate measurements $\vec m_1, \vec m_2, \ldots$ (such as laboratory test results, medication perscripts, infusions, etc.).
        \textbf{(b)} These data elements can be distributed among many input data sources in raw form. From a simple YAML configuration file, \pkgAbbr can extract these data elements from source and compile them into an internal data model consisting of three key dataframes: \lstinline[style=code]{subjects_df} for static data, \lstinline[style=code]{events_df} containing event timestamps and types per subject, and \lstinline[style=code]{dynamic_measurements_df}, storing all observed measurements.
        \textbf{(c)} \pkgAbbr pre-processes these data-frames across several critical axes, doing so efficiently through the use of the Polars library~\cite{polars_software}.
        \textbf{(d)} \pkgAbbr produces a PyTorch dataset which yields batches whose size scales with the number of observed data element per event, not with the input vocabulary size.
        \textbf{(e)} \pkgAbbr provides a default embedding layer capable of embedding these sparse batches efficiently.
    }
    \label{fig:data_e2e}
\end{figure}

\pkgAbbr solves all these challenges in data extraction and pre-processing. Figure~\ref{fig:data_e2e} shows the entire pipeline in visual form; from a flexible, concise, user-friendly configuration file, \pkgAbbr can extract raw data from source, pre-process categorical and numerical data elements, and produce efficient PyTorch datasets for downstream analysis. Below, we detail three key advantages of the \pkgAbbr system, and full technical details of the pipeline and its usage an be found in the SI.

\paragraph{\pkgName is Easy to Use and Flexible}
To extract a machine-learning ready dataset from \emph{any} temporal, structued dataset with \pkgName, a user only needs to specify a modest configuration file detailing the input data sources, desired modalities to extract, and pre-processing pipeline parameters. This configuration file instructs the pipeline how to extract the initial data from its raw source, pre-process the extracted measurements, aggregate the data temporally, and produce deep-learning efficient representation outputs suitable for downstream modeling. A full configuration file for our working example over MIMIC-IV can be found in the appendix.

\paragraph{\pkgName can Efficiently Pre-process Large Datasets}
Through the use of the Polars library~\cite{polars_software} and careful design choices, \pkgAbbr is able to provide a dramatic improvement in data pre-processing speed and output storage costs compared to existing libraries. For our working example, \emph{pre-processing the entire MIMIC-IV dataset (300,000 input patients) with our pipeline takes only approximately 30 minutes, and outputs a dataset that takes only 1.2 GB to store on disk}. While it is difficult to perform a robust comparison across different data extraction pipelines, given they all have differing features and capabilities, anecdotally it is the case that many other pipelines take hours to perform extraction and pre-processing over datasets of similar size, and often occupy much more space on disk. Further details on the computational performance of the pipeline, and illustrations of its performance on other datasets to demonstrate its flexibility can be found in the SI.

\paragraph{\pkgName Provides a Memory-efficient Deep Learning Representation}
Beyond efficient pre-processing of data, \pkgAbbr also provides an efficient deep learning representation and an associated PyTorch Dataset and Data Embedding Layer. This representation has three key benefits compared to existing systems. First, The PyTorch Dataset view of \pkgAbbr datasets is designed for rapid retrieval of a subject's timeseries data, in order, to minimize dataloader iteration time, by storing the entirety of a subject's timeseries data in a single row in the output dataframe, and serializing that row to a plain python list at instantiation. In contrast, existing systems like TemporAI store temporal data as a multi-indexed pandas dataframe object, which requires dynamic slicing within the multi-index levels to retrieve a patient's temporal data.

Second, a patient's data in the deep-learning representation is represented in an extremely sparse format in  \pkgAbbr, so that the memory required to represent a single patient's timeseries scales linearly \emph{only} with the total number of observations that patient has. In contrast, existing systems, such as TemporAI, largely have data storage scale linearly with the \emph{total number of features in the dataset}, in some cases further multiplied by \emph{the maximum number of events any patient has in the dataset}. Given the long-tailed nature of these data, this is an extremely large difference. In our MIMIC-IV working example, for example, the average number of observations per patient event is approximately 80. In contrast, the total number of features in this dataset is 6053 (this includes all measured modalities in Table~\ref{tab:included_modalities}). Therefore, our deep learning batch representation strategy requires only 1.3\% of the required memory of traditional, less sparse solutions.

Beyond these two major improvements, \pkgAbbr also is the only pipeline to fully provide a pre-built PyTorch dataset with a customized embedding layer for our data representation; this means that one can immediately feed any \pkgAbbr dataset into any sequential deep learning pipeline of their choice.

\section{Building GPTs}
\label{sec:model}
\subsection{The Problem} \label{subsec:model_problem}
To produce the generative model outlined in Section~\ref{sec:working_example}, we must satisfy several desiderata:
\begin{enumerate}[nolistsep]
    \item We need to be able model historical dependencies ($p(t_i, \vec x_i | \vec h_{i-1})$), temporal dependencies ($p(\vec x_i | t_i)$), and intra-event dependencies ($p(\vec x_i^{(j+1)} | \vec x_i^{(1)}, \ldots, \vec x_i^{(j-1)})$). In contrast, traditional (sequential) GPTs only need to model historical dependencies.
    \item We need to enable deep, sequential processing of event sequences that can take into account event timestamps to build high-capacity representations.
    \item We need to produce categorical, temporal, and continuous output emission distribution. In contrast, traditional GPTs only need to output categorical emission distributions.
    \item Our model needs to be able to dynamically construct analytically determinable time-dependent features (\eg, age) on the fly during generation.
\end{enumerate}

\subsection{Current State-of-the-art}
There are three categories of existing tools that could help us build models: General-purpose deep learning modeling frameworks with built in model architectures (\eg, TemporAI~\cite{temporai}); other GPT architectures for different problem settings, such as natural language or regularly sampled timeseries; or existing models for autoregressive or generative modeling of EHR or related modalities.

Unfortunately, across all three options, there are no existing tools which sufficiently satisfy the criteria outlined in Section~\ref{subsec:model_problem}. Among existing, general-purpose frameworks, such as TemporAI~\cite{temporai} or OMOP Learn~\cite{omoplearn}, lack any suitable built-in model architectures for this task. Existing GPT architectures, on the other hand, can be very helpful, especially for providing sources to adapt transformer architectures for core multi-layer network structures, but lack key features specific to this modality, such as accounting for unknown output event times and complex inter-event dependencies. Finally, among existing, published generative models for EHR or other similar modalities, all existing architectures have major limitations, as outlined in Table~\ref{tab:existing_models} (see the SI for an expanded tables that includes a broader class of models).

\begin{table}
    \centering \small
\begin{tabular}{lccccccc}
\multirow{2}{*}{Model} & \rot{TTE} & \rot{Continuous Outputs} & \rot{Categorical Outputs} & \rot{Intra-event Dependencies} & \rot{Generative} & \rot{Fine-tuning} & \rot{Zero-shot} \\
\cmidrule(lr){2-5} \cmidrule(lr){6-8}  & \multicolumn{4}{c}{Models} & \multicolumn{3}{c}{Evaluation}\\
\midrule
Doctor AI \cite{DOCTOR_AI_pmlr-v56-Choi16}                            & \chk &      & \chk &      & \chk &      &      \\
CDT \cite{CDT_lee2023clinical}                                        &      &      & \chk &      &      &      & \chk \\
MetaCare++ \cite{METACAREpp_10.1145/3477495.3532020}                  & \chk &      & \chk &      & \chk &      &      \\
CLMBR \cite{CLMBR_steinberg2021language}                              &      &      & \chk &      &      & \chk &      \\
MedTPP \cite{enguehard2020neural}                                     & \chk &      & \chk &      & \chk &      &      \\
MedGPT \cite{MEDGPT_kraljevic2021medgpt}                              & \chk &      & \chk &      & \chk &      &      \\
\midrule
\textbf{\pkgAbbr (Ours)} & \chk & \chk & \chk & \chk & \chk & \chk & \chk \\
\bottomrule \end{tabular}
\caption{A summary of existing foundation models over EHR data, broken down by what modalities of data they model and how they report evaluation results.}    \label{tab:existing_models}
\end{table}

\subsection{Contributions of \pkgName}
The \pkgAbbr library includes a pre-built model architecture that models time-to-event components, continuous and categorical output measurements, handles intra-event dependencies in a user-configurable way, and provides a HuggingFace-compatible API for generating new events with a pre-trained model. We detail these capabilities below, and a full technical reference is in the SI.

\paragraph{\pkgName provides a HuggingFace API compatible interface for building models capable of outputting complex, temporal distributions}
To use one of the pre-defined model architectures in \pkgName, users simply build a specialized configuration object, which is a sub-class of the standard, Hugging Face \lstinline[style=code]{PretrainedConfig}\footnote{\url{https://huggingface.co/docs/transformers/main_classes/configuration}}. Once the config is defined, a model can be instantiated in the same way as any other Hugging Face pre-trained model, either from scratch or from a pre-trained source. In addition to the standard Hugging Face attributes, the \pkgAbbr config also contains several custom attributes specific to our setting, including attributes describing the breakdown of the various measurements and their respective vocabulary sizes, the implied dependency relationships between intra-event covariates, and data embedding configuration parameters (for a full specification, see the Supplement).

All pre-built \pkgAbbr models for generative modeling output a unified model output object. This object maps event generation covariate target names to PyTorch distributions, from which we can easily compute losses or sample new events. The unified \pkgAbbr dataset class and this output object defines the API expected by \pkgAbbr models in further utilities like generative capabilities. Full details can be found in the Supplement.

\paragraph{\pkgName provides a modeling interface for encoding complex, custom intra-event dependencies}
\pkgAbbr comes with two pre-defined model architectures. One (the \lstinline[style=code]{ConditionallyIndependent} model) that matches much of the existing published literature and only supports defining generative models where all intra-event covariates are conditionally independent from one another given the patient's history and one (the \lstinline[style=code]{NestedAttention} model) that supports user-defined dependency relationships between various internal features of $\vec x_i$. Users specify these relationships through the configuration file. For example, one of the models pre-trained in our MIMIC-IV working example relies on the intra-event dependency chain shown in Figure~\ref{fig:MIMIC_IV_dep_graph}. This specification takes the form of a list of lists, and tells the model that it should make the assumption that the internal covariates in a given inner list of $\vec x_i$ are conditionally independent of one another given both (1) the historical representation of all $t_\ell, \vec x_\ell$, $\ell < i$ and (2) the internal covariates in all prior lists in the specified dependency graph. 
Concretely, this means that the model configuration snippet shown in Figure~\ref{fig:MIMIC_IV_dep_graph} defines a generative model such that the predicted type of an event will depend only on the historical representation and the true time of that event, but the predicted values observed for the laboratory tests ordered for this patient will depend on both the time of the event, but also the event type, patient dynamic demographics, and the categorical laboratory test item IDs that are actually ordered for that patient.

We describe the architecture used to model this in full detail in the SI; however, what is more essential for the utility of the \pkgAbbr software library is that this API for specifying internal dependencies, and the associated non-architectural changes required to support it (\eg, enabling generation to respect the multi-stage process implied by the intra-event dependency graph) are general, and can be used across a variety of internal transformer architectures. Thus, this provided model class can serve as a basis for further model development that relies on internal event dependencies in future research without invalidating the benefits of the other tools in \pkgAbbr. All aspects of this process are documented fully in the supplementary information.

\begin{figure}
    \centering
    \scriptsize
    
    \begin{lstlisting}[style=configs,numbers=none]
    structured_event_processing_mode: nested_attention
    measurements_per_dep_graph_level: 
        - ["age", "time_of_day"] # Must start with FUNCTIONAL_TIME_DEPENDENT measurements.
        - [ 
            "event_type", "patientweight", "admission_type",  "admission_location",
            "race", "language", "marital_status", "insurance", "careunit",
            ["lab_itemid", "categorical_only"], ["infusion_itemid", "categorical_only"]
          ]
        - [["lab_itemid", "numerical_only"], ["infusion_itemid", "numerical_only"]]
        - ["procedure_itemid", "medication", "icd_code", "discharge_location"]
    \end{lstlisting}
    
    \caption{A sample dependency graph for our MIMIC-IV working example, shown in YAML syntax. The dependency graph is encoded as a list of lists, such that all measurements within any inner list are constrained to depend only on those measurements in prior lists.}
    \label{fig:MIMIC_IV_dep_graph}
\end{figure}

\paragraph{\pkgName provides seamless generation capabilities for unsupervised, zero-shot evaluation}

A critical use-case for foundation models is the ability to generate new, unseen predictions for the future of an input. To the best of our knowledge, \pkgAbbr is the first pipeline to bring that capability in a systemized fashion to foundation models over event stream data like EHR systems.
To do so, we introduce several key modifications from the traditional Hugging Face generation API, including the ability to support complex emission distributions during generation, dynamic computation of time-dependent input features for newly generated time-points, and generating events in sequence while respecting the causal path. This enables us to use generation out of the box on models over \pkgAbbr datasets. Full details can be found in the SI.

\section{Evaluating Foundation Models}
\label{sec:evaluation}

\subsection{The Problem}
Even after building a foundation model on this non-standard domain, one still has to evaluate that model, both as a concrete representation learning system and in its possible viability as an early stage foundation model. In domains of complex, timeseries of events, like medical record datasets, there are a number of unique challenges that must be addressed in evaluation. We foresee the following metrics to be critical points to assess in early stage foundation models in such domains:
\begin{enumerate}[nolistsep]
    \item Performance as a raw generative model.
    \item Viability as a foundation model
    \item Practical utility vs. current ML systems.
    \item Performance disparities across subject sub-populations
    \item Identifiability of subject-level data from pre-trained model parameters
\end{enumerate}

\subsection{Current State-of-the-art}
Existing models in this space have significant gaps in their evaluation strategies to date. Table~\ref{tab:existing_models}, in addition to summarizing model modalities, also summarizes evaluation methods used. While some techniques, like few-shot fine-tuning analyses, are well represented, others are under-explored.
\emph{In particular, at present there are no existing strategies to reliably assess general, zero-shot performance of GPTs over event stream datasets, a gap which we fill here.}\footnote{The Clinical Decision Transformer (CDT)~\cite{CDT_lee2023clinical} does perform some limited zero-shot evaluation strictly through its reinforcement learning defined policy analysis, not on general tasks.}

\subsection{Contributions of \pkgName}
\pkgAbbr focuses on providing usable, powerful tools to enable rapid evaluation of GPTs over event stream datasets, focused currently on the assessment of a model's generative performance and in its viability as a foundation model in few and zero-shot evaluation tasks. While further efforts on the other evaluation criteria would be highly impactful, given the dearth of existing utilities just on assessing these basic performance metrics, we leave the rest to future work.
In addition to these evaluation utilities, we also provide pre-built scripts for running distributed, Bayesian hyperparameter optimization sweeps over foundation models through weights and biases~\cite{wandb}, to help further standardize that portion of the optimization process.

\paragraph{Assessing Generative Model Performance \& Hyperparameter Tuning}
To make assessing generative performance easier, \pkgAbbr comes pre-equipped with a PyTorch Lightning~\cite{Falcon_PyTorch_Lightning_2019} module for running pre-training models which tracks a variety of metrics to assess generative model performance, including traditional classification and regression metrics over all component internal event covariates. These are logged by default to weights and biases, permitting easy exploration of output model performance. 
These metrics are further tracked and can be assessed visually in weights and biases reports during hyperparameter tuning as well, which is run via a Weights and Biases sweep with a provided default hyperparameter search space. A template report for assessing the output of these sweeps is also provided alongside the code for our working example over MIMIC-IV (accessible in the SI), allowing users to easily design their own dashboards to monitor their model tuning runs.

\paragraph{Assessing Foundation Model Viability via Zero-shot Performance}
While many systems already explore few-shot performance (and \pkgAbbr comes pre-equipped with lightning modules and utilities for performing similar assessments), assessing zero-shot performance across general models has not been previously studied. Enabling zero-shot evaluation for foundation models trained on event stream data requires three things: First, a labeled dataset corresponding to a well-defined task which can be used to evaluate zero-shot produced labels; second, a method to perform unsupervised forecasting of a patient's record given an input window, which \pkgAbbr provides; third, a function to translate generated samples into empirical predictions for the task in question. 

The \pkgAbbr system then provides a PyTorch Lightning utility to take as input a dataset, task, pre-trained model, and labeling function, and perform zero-shot evaluation over a specified data cohort. This system reports performance metrics based on the output of this labeling function over the empirical predictions generated for each patient's input. \emph{This provides any model trained using the \pkgAbbr API the ability to assess zero-shot performance in a meaningful way, out of the box.} For examples of these evaluation utilities on MIMIC-IV, see the SI.

\section{Further Discussion}
\paragraph{Limitations}
While \pkgAbbr provides significant utilities, it also has a number of limitations. Firstly, it lacks some more advanced pre-processing capabilities that other systems support, such as unit conversion, ontological aggregation, or support for a wide variety of optional pre-processing algorithms. Secondly, defining downstream tasks is currently only possible through manual creation / user-defined code, as opposed to via a configuration language. Finally, lastly, while the utilities \pkgAbbr offers do significantly enhance evaluation, even more is needed on that front, including dedicated metrics for fairness and privacy sensitivity, methods to assess embedding space structure, and methods to detect emergence of foundation model capabilities more cost efficiently.

\paragraph{Related Work}
\label{sec:related_work}

As listed in Tables~\ref{tab:existing_pipelines}~\&~\ref{tab:existing_models}, there are a number of existing processing pipelines and models of relevance here. Beyond those already discussed, there are also a number of models that leverage non-generative losses that warrant highlighting, such as contrastive methods like \cite{OCP_agrawal2022leveraging,HIBEHRT_li2021hi,SIPT}. These provide a valuable alternative to generative foundation models in various domains. Additionally, other pipelines also exist in this and adjacent spaces, such as time-series specific pipelines and tabular data embedding methodologies~\cite{KATS,gorishniy2021revisiting,tsfresh}.
GPTs have also been used very successfully in other domains outside of event stream data. These include both examples of foundation models outside of either NLP or event stream data, such as on protein sequences~\cite{ESM} or point clouds~\cite{pointBert}, but also other examples of temporal point processes outside of the typical ``foundation model'' framing~\cite{intensityFree,neuralHawkes}

\section{Conclusion}
In this paper, we introduced \pkgName (\pkgAbbr), a new, open source software library that fills a critical gap in the foundation model literature: namely, technological support for building generative, pre-trained transformers (GPTs) over ``event stream data''---datasets consisting of continuous time sequences of complex events with multiple modalities and internal dependencies. We've shown that \pkgName can reliably address all aspects of the research pipeline for this end, from data pre-processing at scale, model construction in a manner accounting for the unique properties of event stream data, and evaluation across all aspects of the foundation model pipeline. We feel that \pkgAbbr will significantly accelerate, standardize, and improve the space of research into these models and will help identify the best architectures and settings where the strengths of GPTs as seen in other domains can be replicated in new domains.

\begin{ack}
MBAM gratefully acknowledges support by a Berkowitz Postdoctoral Fellowship at Harvard Medical School.
In addition, this work has benefited from the advice of Oleksander Shchur.

The All of Us Research Program is supported by the National Institutes of Health, Office of the Director:
  Regional Medical Centers: 1 OT2 OD026549; 1 OT2 OD026554; 1 OT2 OD026557; 1 OT2 OD026556; 1 OT2 OD026550; 1
  OT2 OD 026552; 1 OT2 OD026553; 1 OT2 OD026548; 1 OT2 OD026551; 1 OT2 OD026555; IAA \#: AOD 16037; Federally Qualified Health Centers: HHSN 263201600085U; Data and Research Center: 5 U2C OD023196; Biobank: 1 U24 OD023121; The Participant Center: U24 OD023176; Participant Technology Systems Center: 1 U24 OD023163; Communications and Engagement: 3 OT2 OD023205; 3 OT2 OD023206; and Community Partners: 1 OT2 OD025277; 3 OT2 OD025315; 1 OT2 OD025337; 1 OT2 OD025276. In addition, the All of Us Research Program would not be possible without the partnership of its participants.
\end{ack}

\bibliographystyle{plain}
\bibliography{refs}

\appendix

\section{Code and Full Technical Documentation of \pkgName}
\pkgName is open source, available at \url{https://github.com/mmcdermott/EventStreamGPT}, with documentation
accessible at \url{https://eventstreamml.readthedocs.io}.

\paragraph{Full Technical Documentation}
Full technical documentation for \pkgAbbr can be found in online at \url{https://eventstreamml.readthedocs.io}. It features:
\begin{enumerate}[nolistsep]
    \item \href{https://eventstreamml.readthedocs.io/en/latest/api/modules.html}{API}: A full API reference of all module classes, functions, configuration files, etc.
    \item \href{https://eventstreamml.readthedocs.io/en/latest/MIMIC_IV_tutorial/index.html}{MIMIC-IV Tutorial}: A full walk-through of our MIMIC-IV example, both in written documentation and in code, with concrete examples of model and evaluation output results over MIMIC-IV, as well as a link to a stand-alone github repository for that example.
    \item \href{https://eventstreamml.readthedocs.io/en/latest/usage.html\#data-pipeline}{Dataset~Pipeline}: A full description of the dataset pipeline and configuration language.
    \item \href{https://eventstreamml.readthedocs.io/en/latest/usage.html\#sample-model-architectures}{Pre-built Architectures}: Full descriptions of our two pre-built model architectures.
\end{enumerate}

\section{Generalizing Beyond MIMIC-IV: Dataset Pipeline on Other Datasets}
The \pkgAbbr pipeline is not specific to MIMIC-IV. We have used it on a number of other datasets, spanning both public and private settings. In Table~\ref{tab:data_in_summary}, we show a summary of the pipeline's usability across three different datasets: the public MIMIC-IV dataset~\cite{mimiciv}, the public All of Us Dataset~\cite{allofus}, which is a large, electronic health record (EHR) dataset consisting of broad spectrum care over a large number of individuals from diverse areas, health statuses, and backgrounds; and the ``HF'' dataset, a private dataset sourced from an ongoing, IRB-approved study regarding the use of machine learning for analysis of heart failure measurements from an academic research institution consisting of labs, vitals, and cardiology derived measures for a large number of patients. We see that the system is able to process large datasets very quickly with only modest configuration inputs from the user with consistent space reductions in all three settings.

\begin{table}
    \centering

    \begin{tabular}{lclrr}\toprule
        \mr{Dataset}            & \mr{Public?} & \mr{Input Format} & \multicolumn{2}{c}{Input Size} \\
                                &              &                   & \# Subjects & Disk (GB)   \\
        \midrule
        MIMIC-IV~\cite{mimiciv} & \chk         & PostgreSQL        & 300k        & 43   \\
        All of Us~\cite{allofus}  & \chk         & Parquet           & 159k        & 11   \\
        HF                      &              & CSV               & 134k        & 32   \\
    \bottomrule\end{tabular}

    \vspace{1em}

    \begin{tabular}{lrrrrrrrrr} \toprule
    \multirow{2}{*}{Dataset} & \multirow{2}{*}{\# Lines} & \multirow{2}{*}{CPUs} & \multirow{2}{*}{Time (min)} & \multirow{2}{*}{Mem (GB)} & \multicolumn{3}{c}{Output} & \multicolumn{2}{c}{Disk (GB)} \\
    \cmidrule(lr){6-8} \cmidrule(lr){9-10}
      &  &  &  &  & Subj. & Events & Meas. & Non-DL & DL\\
    \midrule
    MIMIC-IV    & 153 &  10 & 30.78 ± 1.12 &  65.93 ± 0.14 &  12k &  3M & 222M & 1.19 & 0.49 \\
    All of us$^*$ &  98 &  64 & 62.03 ± 0.31 & 119.36 ± 0.93 & 145k & 28M & 330M & 2.95 & 3.99\\
    HF      &  78 & 128 & 24.96 ± 0.44 & 161.07 ± 0.23 & 116k & 25M & 244M & 3.6  & 1.54\\
    \bottomrule \end{tabular}

    \caption{\textit{Top:} Descriptions of the raw data for each of our four datasets. \textit{Bottom:}
    Statistics on computational cost to produce each dataset. ``\# Lines'' refers to the number of lines in
    the `yaml` configuration file used to produce this dataset. Standard deviations are omitted when they are
    universally $\le 0.0$. Memory was assessed via \texttt{mprof} for MIMIC-IV and HF, but due to technical
    issues it was assessed via a shell loop with the `free` command for All of us.}
    \label{tab:data_in_summary}
\end{table}

\section{Further Details on Existing Models}
There are a variety of existing foundation model attempts over EHR and related datasets. In Tables~\ref{tab:existing_models} and \ref{tab:existing_models_2}, we summarize a variety of existing models over a larger class of options than is shown in the main-body Table 3 (in particular, here we also include Contrastive and Masking based pre-training systems). Even across these broader examples, we see that evaluation is largely limited to fine-tuning evaluations, establishing a clear gap regarding zero-shot evaluation.

\begin{table}
    \centering \small
\begin{tabular}{llccccccccccclrrrr}
\multirow{2}{*}{Style} & \multirow{2}{*}{Model} & \rbox{Primary Care} & \rbox{Hospital} & \rbox{Intensive Care Unit} & \rbox{Claims} & \rbox{Clinical Notes} & \rbox{Diagnoses} & \rbox{Procedures} & \rbox{Medications} & \rbox{Lab Orders} & \rbox{Lab Results} & \rbox{Vitals} & \multirow{2}{*}{``Token''} & \multirow{2}{*}{Patients} & \multirow{2}{*}{T/P} & \multirow{2}{*}{C/T} & \multirow{2}{*}{\# Codes} \\ 
\cmidrule(lr){3-7} \cmidrule(lr){8-13}  &  & \multicolumn{5}{c}{Care Modalities} & \multicolumn{6}{c}{Event Types} &  &  &  &  & \\
\midrule
\multirow{6}{*}{Causal}      & Doctor AI ~\cite{DOCTOR_AI_pmlr-v56-Choi16}                            & \chk &      &      &      &      & \chk & \chk & \chk &      &      &      & Visit &  0.3M & 54.6 & 3.2 &  1.8k\\
                             & CDT ~\cite{CDT_lee2023clinical}                                        &      & \chk &      &      &      &      &      & \chk & \chk & \chk & \chk & Event &  0.0M & 13.6 &     &  0.0k\\
                             & MetaCare++ ~\cite{METACAREpp_10.1145/3477495.3532020}                  &      &      & \chk &      &      & \chk &      &      &      &      &      & Visit &  0.0M &  2.2 & 4.8 &  0.9k\\
                             & CLMBR ~\cite{CLMBR_steinberg2021language}                              &      & \chk &      &      &      & \chk & \chk & \chk & \chk &      &      & 1 day &  3.4M &  7.0 & 5.0 & 21.7k\\
                             & MedTPP ~\cite{enguehard2020neural}                                     &      &      & \chk &      &      &      &      &      &      &      &      & Event &  0.0M &  4.0 &     &  0.1k\\
                             & MedGPT ~\cite{MEDGPT_kraljevic2021medgpt}                              &      &      &      &      & \chk & \chk &      &      &      &      &      & Event &  0.6M &      &     &      \\
\midrule
\multirow{2}{*}{Contrastive} & OCP ~\cite{OCP_agrawal2022leveraging}                                  &      &      &      &      & \chk &      &      &      &      &      &      & Event &  0.1M & 12.0 &     & 30.5k\\
                             & Hi-BEHRT ~\cite{HIBEHRT_li2021hi}                                      & \chk & \chk &      &      &      & \chk & \chk & \chk & \chk &      &      & Visit &  2.8M & 62.0 & 4.5 &  3.7k\\
\midrule
\multirow{10}{*}{Masking}    & Med-BERT ~\cite{MEDBERT_rasmy2021med}                                  & \chk & \chk &      &      &      & \chk &      &      &      &      &      & Visit & 28.5M &      &     & 82.0k\\
                             & BEHRT ~\cite{BEHRT_li2020behrt}                                        & \chk & \chk &      &      &      & \chk &      &      &      &      &      & Visit &  1.6M &      &     &  0.3k\\
                             & Graph-Transformer ~\cite{GRAPH_TRANSFORMER_pellegrini2022unsupervised} &      &      & \chk &      &      &      &      & \chk & \chk & \chk & \chk & 1 hr  &  0.0M & 24.0 &     &      \\
                             & EHR-PT ~\cite{EHR_PT_mcdermott2021comprehensive}                       &      &      & \chk &      &      &      &      & \chk & \chk & \chk & \chk & 1 hr  &  0.1M & 70.6 &     &      \\
                             & RAPT ~\cite{RAPT_10.1145/3447548.3467069}                              &      & \chk &      &      &      &      &      &      & \chk & \chk & \chk & Visit &  0.1M &  6.8 &     &      \\
                             & CEHR-BERT ~\cite{CEHR_BERT_pmlr-v158-pang21a}                          &      & \chk &      &      &      & \chk & \chk & \chk & \chk &      &      & Visit &  2.4M & 14.0 & 5.4 &      \\
                             & GRACE ~\cite{GRACE_10.1145/3534678.3539020}                            &      &      & \chk &      &      &      &      &      & \chk & \chk & \chk & Visit &  0.0M &  6.5 &     &      \\
                             & CMS LDS BERT ~\cite{CMS_LDS_BERT_lahlou2021explainable}                &      &      &      & \chk &      & \chk & \chk &      &      &      &      & Event &  1.2M &      &     & 20.0k\\
                             & MTL GPR ~\cite{MTL_si2021generalized}                                  &      &      &      &      & \chk &      &      &      &      &      &      & 1 hr  &  0.0M & 38.0 &     &      \\
                             & T-BEHRT ~\cite{T_BEHRT_rao2022targeted}                                & \chk & \chk &      &      &      & \chk &      &      &      &      &      & Visit &  6.8M &      &     &      \\
\bottomrule \end{tabular}
\caption{A summary of existing MedLMs. \textit{T/P} refers to ``tokens'' per patient, \textit{C/T} to codes per ``token,'' and \textit{\# Codes} to the number of unique codes present in the dataset. Missing values reflect quantities not reported in the source publication.}
\label{tab:existing_models}
\end{table}

\begin{table}
    \centering \small
\begin{tabular}{llccccccc}
\multirow{2}{*}{Style} & \multirow{2}{*}{Model} & \rot{TTE} & \rot{Continuous Outputs} & \rot{Categorical Outputs} & \rot{Intra-event Dependencies} & \rot{Generative} & \rot{Fine-tuning} & \rot{Zero-shot} \\
\cmidrule(lr){3-6} \cmidrule(lr){7-9}  & & \multicolumn{4}{c}{Models} & \multicolumn{3}{c}{Evaluation}\\
\midrule
\multirow{6}{*}{Causal} &
Doctor AI \cite{DOCTOR_AI_pmlr-v56-Choi16}                            & \chk &      & \chk &      & \chk &      &      \\
& CDT \cite{CDT_lee2023clinical}                                        &      &      & \chk &      &      &      & \chk \\
& MetaCare++ \cite{METACAREpp_10.1145/3477495.3532020}                  & \chk &      & \chk &      & \chk &      &      \\
& CLMBR \cite{CLMBR_steinberg2021language}                              &      &      & \chk &      &      & \chk &      \\
& MedTPP \cite{enguehard2020neural}                                     & \chk &      & \chk &      & \chk &      &      \\
& MedGPT \cite{MEDGPT_kraljevic2021medgpt}                              & \chk &      & \chk &      & \chk &      &      \\
\midrule
\multirow{2}{*}{Contrastive} & OCP ~\cite{OCP_agrawal2022leveraging}                                  &  NA  & NA & NA & NA &  & \chk &  \\
                             & Hi-BEHRT ~\cite{HIBEHRT_li2021hi}                                      &  NA  & NA & NA & NA &  &  &  \\
\midrule
\multirow{10}{*}{Masking}    & Med-BERT ~\cite{MEDBERT_rasmy2021med}                                  &  NA  &      & \chk &      &      & \chk &      \\
                             & BEHRT ~\cite{BEHRT_li2020behrt}                                        &  NA  &      & \chk &      &      & \chk &      \\
                             & Graph-Transformer ~\cite{GRAPH_TRANSFORMER_pellegrini2022unsupervised} &  NA  & \chk & \chk &      &      & \chk &      \\
                             & EHR-PT ~\cite{EHR_PT_mcdermott2021comprehensive}                       &  NA  & \chk & \chk &      &      & \chk &      \\
                             & RAPT ~\cite{RAPT_10.1145/3447548.3467069}                              &  NA  & \chk & \chk &      &      & \chk &      \\
                             & CEHR-BERT ~\cite{CEHR_BERT_pmlr-v158-pang21a}                          &  NA  &      & \chk &      &      & \chk &      \\
                             & GRACE ~\cite{GRACE_10.1145/3534678.3539020}                            &  NA  & \chk & \chk &      &      & \chk &      \\
                             & CMS LDS BERT ~\cite{CMS_LDS_BERT_lahlou2021explainable}                &  NA  &      & \chk &      &      & \chk &      \\
                             & MTL GPR ~\cite{MTL_si2021generalized}                                  &  NA  &      & \chk &      &      & \chk &      \\
                             & T-BEHRT ~\cite{T_BEHRT_rao2022targeted}                                &  NA  &      & \chk &      &      & \chk &      \\ \midrule
Causal & \textbf{\pkgAbbr (Ours)} & \chk & \chk & \chk & \chk & \chk & \chk & \chk \\
\bottomrule \end{tabular}
\caption{A summary of existing foundation models over EHR data, broken down by what modalities of data they model and how they report evaluation results.}    \label{tab:existing_models_2}
\end{table}

\end{document}